\documentclass[11pt]{article}
\ifdefined\AnonymousReview
  \usepackage[review]{acl}
\else
  \usepackage[preprint]{acl}
\fi
\usepackage{times}
\usepackage{latexsym}
\usepackage[T1]{fontenc}
\usepackage[utf8]{inputenc}
\usepackage{microtype}
\usepackage{graphicx}
\usepackage{booktabs}
\usepackage{amsmath,amssymb}
\usepackage{array}
\usepackage{url}
\graphicspath{{figures/}}
\newcommand{\pp}{\,\mathrm{pp}}
\newcommand{\Hold}{\textsc{Hold}}
\newcommand{\Replay}{\textsc{Replay}}

\newcommand{\UniformQwenEightOne}{$-0.75$}
\newcommand{\UniformQwenEightTwo}{$-1.55$}

\hypersetup{
  pdftitle={Third Eye: Beyond Immediate Gains in Recursive Self-Improvement},
  pdfsubject={Paired evaluation of persistent language-model self-updates}}
\ifdefined\AnonymousReview
  \hypersetup{pdfauthor={}}
\else
  \hypersetup{pdfauthor={Sujato Dutta; Padala Indira Sneha; Kypu Mahitha;
    Shashank Vanga; Neeraj Choudhary}}
\fi

\title{Third Eye: Beyond Immediate Gains\\in Recursive Self-Improvement}
\author{\textbf{Sujato Dutta \quad Padala Indira Sneha \quad Kypu Mahitha}\\
\textbf{Shashank Vanga \quad Neeraj Choudhary}\\
Mahindra University}

\begin{document}
\maketitle

\begin{abstract}
Does a useful self-update remain useful after the model updates again?
Third Eye separates immediate performance from the outcome of a further
native update. We evaluate five 3--8B backbones on math and code:
64 fresh runs, three candidates per run, and three continuations per candidate.
Selection and final scoring use disjoint items.
In the balanced Qwen3-4B/Llama-3.2-3B comparison, immediate-selected utility is
near zero, but continuation reduces it by $1.78\pp$
(simultaneous 95\% interval $[0.68,2.88]$).
Candidate-averaged immediate utility is already negative:
selection partly accounts for the apparently neutral first step.
Continuation also lowers utility for SmolLM3-3B and Llama-3.1-8B;
Qwen3-8B remains inconclusive.
Target and out-of-distribution losses dominate retention changes.
Matched recipe and recipient--donor controls identify neither a reliable
mitigation nor a single mechanism.
The results support evaluating persistent self-updates as adaptation
transitions, with independent scoring and a no-update baseline, rather
than interpreting immediate performance as their full value.
\end{abstract}

\begin{figure*}[t]
\centering\includegraphics[width=\textwidth]{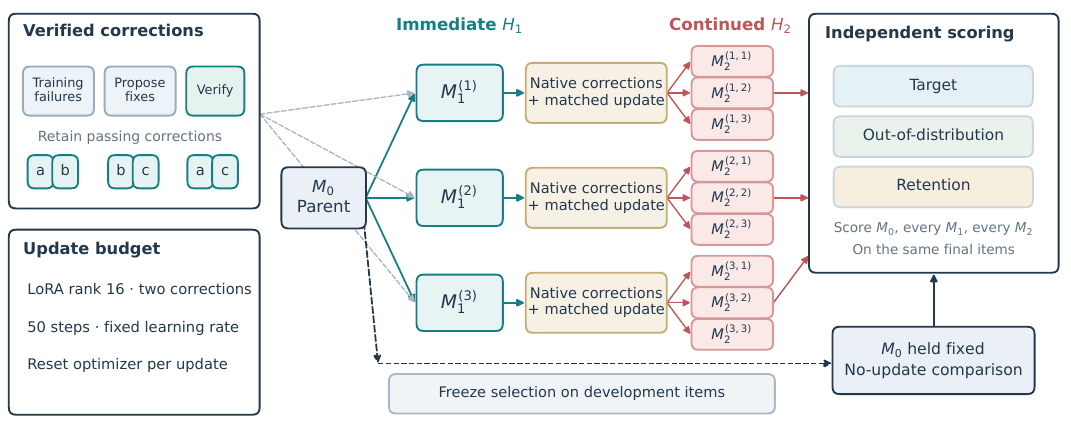}
\caption{Third Eye follows every candidate through another self-update.
Development items select candidates; independent items score both horizons
against the unchanged parent.}
\label{fig:overview}
\end{figure*}

\section{Introduction}

Language models increasingly generate the data used to update their own
parameters \citep{star,selfimprove,selfreward,seal}.
In a recursive loop, an update changes both the model's current answers and
the model that will supply and learn from its next training batch.
An immediate evaluation measures only the current checkpoint.

Verification addresses a different question. Answer checks and execution tests
can reject incorrect corrections \citep{verifiers,process,evalplus}, but passing
examples do not establish the broader value of training on them.
That value may also change after another update.

SEAL learns self-edits using post-adaptation feedback \citep{seal}, while
modification screening checks protected behavior \citep{harness}.
We examine the next native update, where persistent parameter changes also
affect the generation of new training data.

Third Eye measures both horizons (Figure~\ref{fig:overview}).
It constructs three equal-budget LoRA candidates from a common parent, then
follows each through three native continuations.
Development items determine candidate selection; independent items score all
checkpoints against the unchanged parent.
The five-model, two-task study contains 64 fresh runs, 192 candidates, and
576 continuations.

In the balanced Qwen3-4B/Llama-3.2-3B comparison, immediate-selected utility is
$+0.02\pp$ and continued utility is $-1.76\pp$.
Their paired difference has an adjusted interval excluding zero.
However, candidate-averaged immediate utility is already negative:
selection partly explains the neutral first-step mean.
SmolLM3-3B and Llama-3.1-8B instead show immediate damage followed by further
decline; Qwen3-8B's continuation effect remains unresolved.
The evidence concerns different assessments across horizons, with substantial
model dependence.

We contribute:
\begin{enumerate}
\item A paired protocol with independent scoring, repeated continuations, and an explicit no-update baseline.
\item Five-model evidence separating selection effects and
target/OOD/retention changes across horizons.
\item Matched recipe and recipient--donor controls that bound explanations and proposed mitigations.
\end{enumerate}

\section{Measuring a Self-Update Beyond Its First Step}
\label{sec:method}

\subsection{Verified data and matched branches}

A parent model $M_0$ first attempts training problems. Failed problems receive
up to eight correction attempts each. Only solutions that pass the task
verifier enter the correction pool. Math corrections are checked against
answers; code corrections must pass execution tests in an isolated sandbox.
We construct $K=3$ compositionally distinct batches $B_k$, each containing two
different problems. Batches may overlap with one another.
This preserves candidate diversity without requiring six different
verified corrections.

Let $A$ denote the fixed update procedure. Candidate $k$ is
\begin{equation}
M_1^{(k)}=A(M_0,B_k;\xi_k).
\end{equation}
The parent is restored before each candidate is trained.
For each draw $d\in\{1,2,3\}$, candidate $M_1^{(k)}$ generates a native
correction batch $C_{k,d}$ and undergoes one further update:
\begin{equation}
M_2^{(k,d)}=A(M_1^{(k)},C_{k,d};\zeta_{k,d}).
\end{equation}
All candidates continue, including those not chosen by the immediate selector.
Continuation draws vary the correction/optimization procedure jointly; they
do not separately randomize each source of variation.

Updates use LoRA \citep{lora}, rank 16 and scaling 32, on attention projections.
Each full update has 50 AdamW steps, learning rate $10^{-4}$, microbatch size
one, and gradient accumulation eight. Thus two corrections receive 400 example
presentations per update. Optimizer state is reset between updates, while the
candidate's parameter changes persist into its continuation.
If a native correction pool is exhausted without supplying a full batch,
the branch is terminal: $M_2=M_1$, with no continuation training.
None of the 576 fresh continuation draws was terminal.

\subsection{Immediate and continued utility}

For role $r\in\{\mathrm{target},\mathrm{OOD},\mathrm{retention}\}$, let
$a_r(M)$ be accuracy on a fixed item set. Utility is the equal-role mean
change from the parent:
\begin{equation}
U(M;M_0)=\frac{1}{3}\sum_r\big[a_r(M)-a_r(M_0)\big].
\label{eq:utility}
\end{equation}
We report percentage points, multiplying this quantity by 100.
Equal role weights prevent the larger retention set from dominating.
They remain a specified objective, rather than a universal measure of quality.

The immediate selector maximizes development utility. For run $p$, let
$w_{pk}$ assign equal mass to tied best candidates and zero elsewhere.
On independent scoring items, its two endpoints are
\begin{align}
G_{1p}&=\sum_k w_{pk}U(M_1^{(k)};M_0),\\
G_{2p}&=\sum_k w_{pk}\frac{1}{3}\sum_{d=1}^{3}
U(M_2^{(k,d)};M_0).
\end{align}
Their paired difference, $\Delta_p=G_{2p}-G_{1p}$, measures the increment
from continuation. Both endpoints use the same parent reference.
Here $H_2$ means an observed two-update outcome, not a prediction.

We also report uniform-candidate endpoints, replacing $w_{pk}$ by $1/3$.
This separates the behavior of the update procedure from the effect of
selecting a candidate on development performance.
\Hold{} retains $M_0$ and has zero utility by definition.
\Replay{} admits the immediate-selected candidate only if its development
target gain is positive and both OOD and retention gains are nonnegative.
These actions are recorded before final scoring. Holding does not consume an
equal training budget; it is the decision alternative to updating.

\begin{table}[t]
\centering\small
\setlength{\tabcolsep}{3.5pt}
\begin{tabular}{lll}
\toprule
Role & Math & Code\\
\midrule
Training & GSM8K: 1,024 & MBPP: 306\\
Dev.\ target & GSM8K: 64 & MBPP: 64\\
Dev.\ OOD & MATH: 64 & MBPP val.: 64\\
Final target & GSM8K: 128 & MBPP test: 128\\
Final OOD & MATH-500: 128 & HumanEval: 128\\
Retention & \multicolumn{2}{c}{MMLU: 256 in each suite}\\
\bottomrule
\end{tabular}
\caption{Fixed evaluation suites. Development and final items are disjoint.
Retention is measured on separate development and final subsets.}
\label{tab:data}
\end{table}

\subsection{Models, tasks, and experimental units}

We evaluate Qwen3-4B and Qwen3-8B \citep{qwen3},
Llama-3.2-3B-Instruct \citep{llama32card},
Llama-3.1-8B-Instruct \citep{llama3}, and SmolLM3-3B \citep{smollm3}.
The three smaller backbones have eight fresh runs per task; each 8B backbone
has four. This gives 64 runs, balanced between math and code within every
model. A run repeats the adaptation experiment from the same released
checkpoint with a different random seed; it is not an independently
pretrained model.

Table~\ref{tab:data} specifies the data. GSM8K \citep{verifiers} and
MBPP \citep{mbpp} provide training failures and target evaluation.
MATH \citep{math} and HumanEval \citep{humaneval} provide final OOD
problems; MMLU \citep{mmlu} measures retention.
Code's development OOD set is MBPP validation, whereas final OOD scoring uses
HumanEval. This domain change is relevant when interpreting admission.
Eight HumanEval problems used for timing calibration are excluded from
final scoring.

Every run scores 13 checkpoints: the parent, three candidates, and nine
continuations. The study therefore contains 832 checkpoint evaluations.
Checkpoints and draws are repeated measurements within a run, not additional
independent runs. Hash-locked manifests, templates, actions, and checkpoint
identities link these measurements. We freeze the date in model templates,
preventing a changing system date from changing prompts across checkpoints.
Appendix~\ref{app:implementation} documents settings and exact revisions.

\subsection{Uncertainty and declared comparisons}
\label{sec:inference}

We use 5,000 bootstrap replicates, resampling parent runs within model/task
strata and items within evaluation roles. Checkpoints sharing an exact item
pool share its item resample. This retains pairing and avoids treating reuse
of the same test questions as independent evidence
\citep{variance,dror,berg}.
The three native continuation draws are averaged within a candidate before
resampling; they are not separately resampled. Intervals are consequently
conditional on these finite draws and the frozen selector.

Declared tests and descriptive comparisons are kept separate.
For Qwen3-4B/Llama-3.2-3B, the primary is the model difference in $G_2$,
with a joint 95\% interval. The pooled continuation contrast belongs to a
six-contrast secondary family, with simultaneous 95\% intervals and
Holm-adjusted centered-bootstrap $p$-values \citep{holm}.
SmolLM3's declared primary is its paired continuation contrast, with a
95\% interval. Each 8B model has the same endpoint with a 97.5\% interval,
giving joint 95\% coverage for the two-model test family.
Each declaration was frozen before its corresponding outcomes.
All five-model plots and role/task decompositions are descriptive;
they do not create a new pooled confirmatory primary.

\section{What Changes After Another Update?}
\label{sec:results}

\begin{figure*}[t]
\centering\includegraphics[width=\textwidth]{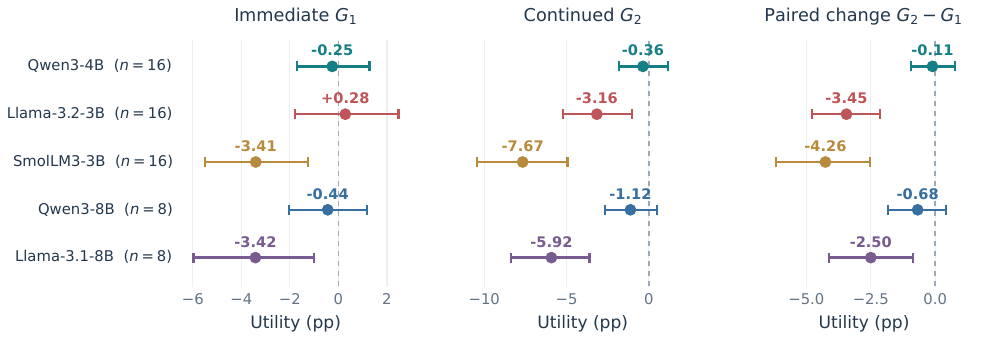}
\caption{Immediate and continued utility across five models.
Bars are descriptive joint run/item 95\% intervals; adjusted intervals for
declared contrasts appear in Table~\ref{tab:contrasts}. The paired change
uses within-run differences, rather than differences between interval endpoints.}
\label{fig:horizons}
\end{figure*}

\subsection{A neutral immediate mean can precede a decline}

In the 32-run Qwen3-4B/Llama-3.2-3B comparison, immediate-selected utility
averages $+0.016\pp$ (95\% interval $[-1.319,1.326]$).
After continuation, utility is $-1.760\pp$ ($[-3.105,-0.420]$).
The paired decrement is $-1.776\pp$, with simultaneous interval
$[-2.876,-0.679]$ and Holm-adjusted $p=0.0012$.
The same selected candidates therefore receive materially different
assessments at the two horizons.

The near-zero immediate mean is not an equivalence test: its interval permits
both benefit and harm. It also combines different model responses.
Llama-3.2-3B has immediate utility $+0.285\pp$ and continued utility
$-3.160\pp$; Qwen3-4B has $-0.252\pp$ and $-0.359\pp$.
Most of the pooled continuation decrement comes from Llama.
The predeclared Qwen-minus-Llama $G_2$ contrast is $+2.801\pp$
($[0.266,5.329]$).
This is evidence of different outcomes under a common recipe, not evidence
that Qwen adaptation is beneficial: its mean remains slightly negative.

\subsection{Selection accounts for part of the first-step assessment}

Uniform candidate weights change the interpretation of the first step
(Table~\ref{tab:uniform}). Across the same 32 runs, candidate-averaged
immediate utility is $-0.686\pp$, rather than the selector's $+0.016\pp$.
Candidate-averaged continued utility is $-2.059\pp$.
Immediate selection thus raises both endpoints, especially the first,
but does not eliminate the continuation decrement.

This control limits the claim that immediate evaluation simply misses all
update damage. Some damage is already visible before selection, and
SmolLM3 and Llama-3.1-8B have negative immediate outcomes even after selection.
What the paired design establishes is that the immediate endpoint alone
does not describe the full update procedure.

\begin{table}[t]
\centering\small
\setlength{\tabcolsep}{4pt}
\begin{tabular}{lrrrr}
\toprule
& \multicolumn{2}{c}{Selected} & \multicolumn{2}{c}{Uniform}\\
\cmidrule(lr){2-3}\cmidrule(lr){4-5}
Model & $G_1$ & $G_2$ & $G_1$ & $G_2$\\
\midrule
Qwen3-4B & $-0.25$ & $-0.36$ & $-0.50$ & $-0.41$\\
Llama-3.2-3B & $+0.28$ & $-3.16$ & $-0.87$ & $-3.71$\\
SmolLM3-3B & $-3.41$ & $-7.67$ & $-3.65$ & $-8.02$\\
Qwen3-8B & $-0.44$ & $-1.12$ & \UniformQwenEightOne & \UniformQwenEightTwo\\
Llama-3.1-8B & $-3.42$ & $-5.92$ & $-4.20$ & $-7.56$\\
\bottomrule
\end{tabular}
\caption{Mean utility in percentage points. Uniform weights average all
three candidates, removing development-based candidate selection.}
\label{tab:uniform}
\end{table}

\subsection{Continuation effects differ across backbones}

Figure~\ref{fig:horizons} presents the full model matrix.
SmolLM3-3B declines by $4.262\pp$ after continuation, with its declared
95\% interval $[-6.163,-2.581]$.
Llama-3.1-8B declines by $2.501\pp$, with adjusted 97.5\% interval
$[-4.411,-0.583]$.
Neither replicates the near-neutral first step: their immediate means are
$-3.406\pp$ and $-3.418\pp$, respectively.
They support an additional continuation cost under this recipe, while
showing that delayed loss need not be hidden initially.

Qwen3-8B's mean continuation change is $-0.681\pp$, with adjusted interval
$[-1.914,0.629]$. This result is inconclusive.
Its interval admits both a meaningful decline and a modest improvement.
The five-model pattern is therefore neither uniform degradation nor a
monotonic size effect. Only Qwen3-8B was executed on A100 hardware; the
remaining fresh runs used GH200 hardware. Model size, family, training
history, and runtime are not independently randomized.

\begin{table}[t]
\centering\small
\setlength{\tabcolsep}{3pt}
\begin{tabular}{lrl}
\toprule
Declared contrast & Mean & Interval\\
\midrule
Qwen4B $-$ Llama3B, $G_2$ & $+2.80$ & $[+0.27,+5.33]$\\
Pooled $\Delta$, these models & $-1.78$ & $[-2.88,-0.68]$\\
SmolLM3-3B, $\Delta$ & $-4.26$ & $[-6.16,-2.58]$\\
Qwen3-8B, $\Delta$ & $-0.68$ & $[-1.91,+0.63]$\\
Llama-3.1-8B, $\Delta$ & $-2.50$ & $[-4.41,-0.58]$\\
\bottomrule
\end{tabular}
\caption{Declared contrasts, in percentage points. Coverage is 95\% for
the first and third rows, simultaneous 95\% for the second, and 97.5\%
per model for the two 8B rows. These are separate declared test families.}
\label{tab:contrasts}
\end{table}

\subsection{Target and OOD loss dominate retention loss}

\begin{figure}[t]
\centering\includegraphics[width=\columnwidth]{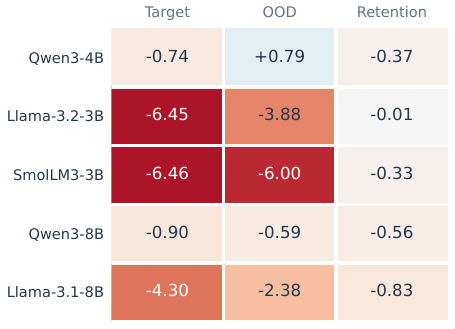}
\caption{Role-wise continuation changes, in percentage points.
Target and OOD losses dominate in the backbones with the largest mean
declines. Colors show descriptive means, not significance.}
\label{fig:heads}
\end{figure}

Figure~\ref{fig:heads} decomposes the continuation increment by role.
For Llama-3.2-3B, target accuracy falls by $6.45\pp$ and OOD accuracy by
$3.88\pp$, while retention changes by only $-0.01\pp$.
SmolLM3-3B shows target/OOD changes of $-6.46/-6.00\pp$, versus
$-0.33\pp$ in retention.
Llama-3.1-8B shows the same ordering, with
$-4.30/-2.38/-0.83\pp$ for target/OOD/retention.
Thus the observed loss is not adequately described as forgetting unrelated
MMLU knowledge. It also concerns the capabilities adaptation is meant to serve.

Qwen3-4B differs: target declines by $0.74\pp$, OOD improves by
$0.79\pp$, and retention declines by $0.37\pp$.
These are descriptive role means, not three independently established effects.
An exploratory model-by-task interaction in the 32-run comparison is
$+2.07\pp$ for the continuation contrast, with 95\% interval
$[-0.98,5.11]$; it does not resolve a task-specific explanation.
Appendix~\ref{app:decomposition} gives both horizons for every role.

Formatting also changes. On GSM8K, strict numeric-extraction failures rise
from 4.49\% at $H_1$ to 10.19\% at $H_2$ for Llama-3.2-3B, and from
4.69\% to 15.12\% for SmolLM3.
These are observable failure modes within the measured loss.
They do not establish how much loss reflects formatting rather than
problem-solving ability. We preserve the original scoring and do not repair
answers after seeing outcomes.
Appendix~\ref{app:format} reports syntax, extraction, and length diagnostics.

\begin{figure}[t]
\centering\includegraphics[width=\columnwidth]{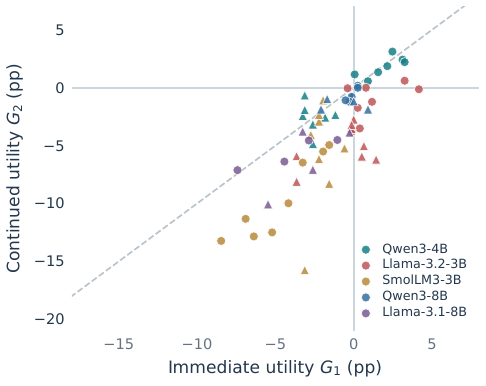}
\caption{Each point is one fresh run; circles denote code and triangles
math. The diagonal marks unchanged utility after continuation.
Observed signs are finite-item estimates, not noise-corrected labels.}
\label{fig:parents}
\end{figure}

Run-level signs also differ (Figure~\ref{fig:parents}).
Seven of nine Llama-3.2-3B runs with positive measured $G_1$ have negative
measured $G_2$; none of Qwen3-4B's eight positive-$G_1$ runs does.
SmolLM3 and Llama-3.1-8B have no positive-$G_1$ runs in this sample.
These observed signs describe finite-item outcomes, not noise-corrected
individual effects.

\section{Controls on the Update Procedure}
\label{sec:controls}

The main comparison couples the receiving parameters, the corrections they
generate, and subsequent optimization. We examine these factors in 16 paired
follow-up parents, four per Qwen3-4B/Llama-3.2-3B task stream.
They reuse inspected parents and are not additional fresh replications.
One fixed candidate branch and two draws per parent yield 272 scored cells.
Native baselines and controls are re-scored on the same A100 runtime.
The nine declared control contrasts use joint run/item resampling,
simultaneous 95\% intervals, and Holm correction.

\subsection{Fewer steps and pooled corrections do not give a resolved rescue}

To test the intensity of the two-example recipe, we compare native
50-step continuations with 10 steps, 25 steps, and the union of verified
corrections from that candidate's archived continuation batches.
The alternatives change the receiving model's continuation recipe; they do
not redefine the main study's training budget.

\begin{figure}[t]
\centering\includegraphics[width=\columnwidth]{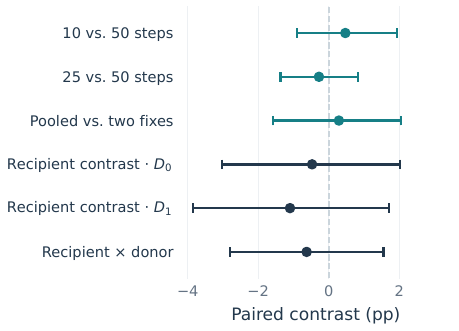}
\caption{Matched recipe and recipient--donor contrasts.
Bars are simultaneous 95\% intervals for the nine-control family.
No displayed intervention resolves an improvement or a single mechanism.}
\label{fig:controls}
\end{figure}

Relative to 50 steps, the 10-step contrast is $+0.47\pp$
($[-0.90,1.94]$), and the 25-step contrast is $-0.28\pp$
($[-1.37,0.83]$).
Pooling yields $+0.29\pp$ ($[-1.57,2.04]$).
All intervals cross zero (Figure~\ref{fig:controls}).
These controls neither establish a mitigation nor prove recipe invariance:
they remain compatible with useful improvements.
The native continuation increment in this smaller subset is
$-1.20\pp$ ($[-3.06,0.55]$), itself inconclusive.

\subsection{Crossed recipients and donors leave the mechanism open}

We apply correction batches from the unchanged parent ($D_0$) and the
first-step model ($D_1$) to both receiving parameter states.
Comparing receivers on the same donor data tests whether receiving state
changes the update increment.
The recipient contrasts are $-0.47\pp$ for $D_0$
($[-3.02,2.03]$) and $-1.10\pp$ for $D_1$
($[-3.85,1.72]$).
Their interaction is $-0.63\pp$ ($[-2.79,1.56]$).
The intervals do not isolate parameter interference, endogenous data change,
or their interaction.
Applying the same native recipe directly to $M_0$ gives
$-0.44\pp$ ($[-2.40,1.36]$).
Thus these controls do not establish that the additional loss requires the
first update, rather than another application of this small-batch recipe.

The donor comparison also has a specific limit: $D_0$ was harvested on the
control runtime, while $D_1$ was archived. Its source contrast includes
generation/runtime differences alongside correction composition.
The crossed design is informative about these matched updates, but it does
not justify a causal decomposition of all native continuation loss.

\subsection{Reference solutions expose another recipe dependence}

A matched reference-solution arm replaces verified self-generated solutions
with gold solutions for the same training problems.
Under the same 50-step recipe, it is $11.56\pp$ worse than the self-generated
arm (simultaneous interval $[-14.45,-8.64]$).
This large difference is not evidence that correct reference solutions are
intrinsically inferior.

A saved-output audit confirms identical receiving states, training seeds,
problem identities, and step budgets.
Yet gold solutions contain only 41--61\% as many assistant-label tokens per
example. Their resulting completions are shorter, and code syntax-failure
rates increase. For Llama code, target syntax failures rise from 1.66\%
to 18.55\%; for Qwen, from 0.98\% to 4.88\%.
Equal example presentations therefore do not match text style or label-token
exposure. These observations motivate caution about data substitution,
rather than a claim that syntax or exposure mediates the entire utility loss.

\section{Implications for Evaluating Self-Improvement}

\paragraph{Measure transitions as well as checkpoints.}
The result is not that immediate evaluation is useless.
It detects substantial first-step damage in several backbones, and candidate
selection improves the immediate mean.
Its limit is that a current score omits the subsequent update transition.
A useful evaluation should report $G_1$, $G_2$, and their paired difference,
with an unchanged parent as a decision baseline.
The role breakdown matters: a retention-only gate would overlook much of the
target/OOD loss observed here.

\paragraph{Abstention is a baseline, not a demonstrated solution.}
In the 32-run comparison, \Replay{} commits 21.9\% of the selected candidate
mass and yields $-0.27\pp$ ($[-0.80,0.21]$) relative to holding.
It does not establish improvement over \Hold{}.
A fixed development-derived Qwen-commit/Llama-hold rule likewise fails to
outperform holding or replay. Better performance than forced updating is
not sufficient evidence of a useful admission policy.

\paragraph{Selection and uncertainty require separate controls.}
Selecting and scoring the same outcome creates optimism
\citep{cawley}. An eight-parent diagnostic shows same-outcome selection
optimism increasing from zero at $K=1$ to $2.09\pp$ at $K=3$
(Appendix~\ref{app:selection}).
Our main study instead freezes selection on separate items.
Uncertainty still depends on shared final questions: for the Qwen--Llama
primary, the run-only interval is $[1.85,3.78]\pp$, compared with the
joint interval $[0.27,5.33]\pp$.
Repeated checkpoints cannot substitute for broader evaluation support.

\paragraph{Prediction remains a distinct empirical claim.}
Training-response forecasting \citep{tuneahead,microinterventions} and
modification screening \citep{rehearse,predictmodify} address useful but
different decisions. The present results establish a measurement contrast;
they do not establish that a forecaster can exploit it.
Earlier held-out forecasting and admission diagnostics did not yield a
validated advantage and are not the basis of the five-model claims.
No learned predictor is fitted to, or selected using, these fresh outcomes.

\section{Related Work}

\paragraph{Self-generated adaptation and agent improvement.}
STaR \citep{star} and self-improvement with generated reasoning
\citep{selfimprove} show how model outputs can become training data.
Self-Rewarding Language Models \citep{selfreward} extend this idea to
preference supervision. SEAL \citep{seal} learns to generate self-adaptation
data and directives using downstream performance after training.
Our study does not introduce self-generated parameter adaptation; it measures
how a candidate's immediate value differs from its value after another
native update.
Self-Refine \citep{selfrefine}, Reflexion \citep{reflexion}, and analyses of
self-correction \citep{selfcorrect} concern refinement and feedback that
need not persist as weight changes.
The Darwin G\"odel Machine \citep{dgm} evolves agent programs.
These settings motivate recursive evaluation but differ from the fixed
LoRA branches studied here.

\paragraph{Predicting and screening modifications.}
TuneAhead \citep{tuneahead} predicts full fine-tuning outcomes from data
descriptors and short probes. Target-independent micro-interventions
\citep{microinterventions} characterize training response across models.
Rehearse \citep{rehearse} and Predict Before You Modify
\citep{predictmodify} use prior modification outcomes to guide agent changes.
The Self-Healing Harness \citep{harness} screens persistent instruction
changes against triggering failures and protected behavior.
Thus beneficial local edits with collateral costs are already documented.
Our distinction is the paired immediate/continued evaluation of persistent
weight updates, with native correction regeneration and independent scoring.

\paragraph{Forgetting and adaptation geometry.}
Continual-learning methods constrain important parameters or replay past
experience \citep{ewc,si,gem}. Forgetting during language-model fine-tuning
\citep{forgetting}, distorted pretrained features \citep{featureshift},
and LoRA-specific intruder dimensions \citep{intruders} offer plausible
explanations for adaptation failures.
LoRA \citep{lora} and QLoRA \citep{qlora} make parameter-efficient experiments
accessible, but efficiency does not establish update quality.
We do not measure spectral mediation or test a continual-learning remedy;
our recipient--donor and recipe controls leave the mechanism unresolved.

\paragraph{Verification and credible evaluation.}
Outcome verification \citep{verifiers}, process supervision \citep{process},
and stronger code testing \citep{evalplus} improve supervision.
They answer different questions from the broader effect of a parameter update.
Evaluation reproducibility \citep{reproducibility}, model-selection bias
\citep{cawley}, benchmark variation \citep{variance}, and paired statistical
testing \citep{dror,berg} motivate our independent scoring and shared-item
resampling. Third Eye applies these safeguards to the adaptation transition,
where candidates and descendants reuse both ancestors and evaluation items.

\section{Conclusion}

Third Eye evaluates a verified self-update immediately and after a further
native update. The five-model study shows why these endpoints should be
reported together: continuation can worsen utility substantially, while
the size and visibility of that change depend on the backbone.
Candidate selection explains part of an apparently neutral first step, and
target/OOD losses can dominate retention changes.
The matched controls provide neither a universal mechanism nor a reliable
repair. The practical contribution is a controlled way to measure update
consequences before interpreting immediate gains as recursive improvement.

\section*{Limitations}

The evidence covers five 3--8B backbones, two tasks, two-example LoRA updates,
and $H_2$. Transfer to larger batches, full fine-tuning, or longer loops is
untested. Fifty steps on two corrections may over-specialize; the controls
cannot rule out better recipes.
Eight runs per 8B model provide less precision.
Hardware, checkpoint training history, and parameter count are not isolated
causal factors.

Fixed item sets offer finite coverage and may overlap pretraining data.
Intervals condition on three averaged continuations, the development selector,
and the equal-role objective. Code's development/final OOD domains differ.
Observed sign changes are noisy estimates.
The 16-parent controls reuse inspected parents; their reference-source audit
does not establish a mediator.

\section*{Ethical Considerations}

We use public math, code, and general-knowledge benchmarks, with no human
participants or private personal data.
Generated code runs in resource-limited isolated sandboxes.
Models and datasets retain their respective license requirements.
Benchmark verification cannot certify unattended self-modification.
We report adverse and unresolved outcomes, and do not interpret
nonsignificance as evidence of safety.

\bibliography{references}

\clearpage
\appendix
\section{Implementation and Frozen Settings}
\label{app:implementation}

\paragraph{Study coverage.}
Table~\ref{tab:coverage} summarizes the model/task matrix.
All models use unquantized bfloat16 arithmetic.

\par\medskip\noindent\begin{minipage}{\columnwidth}
\centering\small
\setlength{\tabcolsep}{3pt}
\begin{tabular}{lcc}
\toprule
Model & Code / math runs & GPU\\
\midrule
Qwen3-4B & 8 / 8 & GH200\\
Llama-3.2-3B & 8 / 8 & GH200\\
SmolLM3-3B & 8 / 8 & GH200\\
Qwen3-8B & 4 / 4 & A100\\
Llama-3.1-8B & 4 / 4 & GH200\\
\bottomrule
\end{tabular}
\captionof{table}{Study coverage. Counts denote fresh adaptation runs from a fixed
released checkpoint, not independent pretraining runs.}
\label{tab:coverage}
\end{minipage}\par\medskip

\paragraph{Optimization.}
We use unquantized bfloat16 models with attention query, key, value, and output
LoRA projections; rank 16, alpha 32, and zero adapter dropout.
The learning rate is $10^{-4}$, with 50 AdamW steps, microbatch one,
gradient accumulation eight, and maximum training sequence length 1,024.
The two-example batch is repeatedly presented, giving 400 presentations.
The optimizer is reinitialized between updates. Parent restoration and
adapter hashes prevent candidate branches from inheriting one another.
The exact optimizer and tokenizer configurations accompany the experiment
manifests.

\paragraph{Generation and scoring.}
Generation is capped at 512 new tokens. Evaluation uses greedy pass@1,
with conditional-likelihood scoring for multiple-choice retention.
Generation batches are 32 for greedy decoding and 16 for sampling.
Reasoning-capable checkpoints use non-thinking mode.
Training failure harvesting scans at most 1,024 GSM8K or 306 MBPP problems,
with eight correction attempts per failed problem. Only verified examples
are retained.
Scoring uses separate immutable development and final manifests.
Prompt and item hashes are checked for reuse; this does not audit external
pretraining corpora.

\paragraph{Scarce corrections.}
Candidate batches must differ in composition and contain no repeated problem
within a batch. Sharing a correction across candidate batches is permitted.
Candidates have equal batch size and optimizer budget.
A continuation with fewer than two verified corrections after its fixed
harvest budget preserves the incoming adapter exactly and records
\texttt{continuation\_available=0}. No synthetic repetition is used to force
an update. All fresh study draws had available continuations.

\paragraph{Frozen prompts.}
A template's automatically inserted date can otherwise change between model
evaluations. We fix the date string to \texttt{11 Oct 2026} and bind template
and prompt hashes to the scoring manifest. Historical runs with changing
templates are excluded from the fresh five-model estimates.

\paragraph{Checkpoint revisions.}
Qwen3-8B is pinned to
\texttt{b968826d9c46}.
SmolLM3-3B is pinned to
\texttt{a07cc9a04f16}.
The Llama-3.1-8B-Instruct checkpoint uses the
NousResearch mirror at revision
\texttt{0e9e39f249a1},
with recorded upstream revision
\texttt{d10aef7999a2}.
These are abbreviated revision identifiers. Full revisions and file hashes are included in the
machine-readable configurations.

\section{Role and Task Decomposition}
\label{app:decomposition}

The tables use compact model labels: Qwen4B and Qwen8B denote Qwen3;
Llama3B denotes Llama-3.2-3B, Llama8B denotes Llama-3.1-8B, and
Smol3B denotes SmolLM3-3B.

The decompositions are exploratory. Their ordinary 95\% intervals are not
adjusted for every displayed role, task, and model comparison.
The model/task interaction reported in the main text is a difference of
paired continuation changes, not an experimentally identified causal effect
of belonging to a model family.

\paragraph{Target accuracy.} The target head evaluates the task distribution used to harvest corrections.
\par\medskip\noindent\begin{minipage}{\columnwidth}
\centering\small\setlength{\tabcolsep}{2pt}
\begin{tabular}{lrrrr}
\toprule
Model & $G_1$ & $G_2$ & $\Delta$ & 95\% CI\\
\midrule
Qwen4B & $+1.64$ & $+0.90$ & $-0.74$ & $[-2.73,+1.24]$\\
Llama3B & $+0.68$ & $-5.76$ & $-6.45$ & $[-9.21,-3.88]$\\
Smol3B & $-5.00$ & $-11.47$ & $-6.46$ & $[-11.30,-2.48]$\\
Qwen8B & $-0.34$ & $-1.24$ & $-0.90$ & $[-3.45,+1.50]$\\
Llama8B & $-4.69$ & $-8.98$ & $-4.30$ & $[-7.98,-0.75]$\\
\bottomrule
\end{tabular}
\captionof{table}{Target accuracy endpoints and paired changes (pp). Intervals are descriptive joint run/item intervals.}
\label{tab:role_target_test}
\end{minipage}\par\medskip
\paragraph{Out-of-distribution accuracy.} The OOD head tests transfer to MATH-500 or HumanEval.
\par\medskip\noindent\begin{minipage}{\columnwidth}
\centering\small\setlength{\tabcolsep}{2pt}
\begin{tabular}{lrrrr}
\toprule
Model & $G_1$ & $G_2$ & $\Delta$ & 95\% CI\\
\midrule
Qwen4B & $-1.73$ & $-0.94$ & $+0.79$ & $[-0.93,+2.58]$\\
Llama3B & $-0.51$ & $-4.39$ & $-3.88$ & $[-6.56,-1.27]$\\
Smol3B & $-5.30$ & $-11.30$ & $-6.00$ & $[-8.71,-3.42]$\\
Qwen8B & $+0.05$ & $-0.54$ & $-0.59$ & $[-2.93,+2.02]$\\
Llama8B & $-6.74$ & $-9.11$ & $-2.38$ & $[-5.27,+0.52]$\\
\bottomrule
\end{tabular}
\captionof{table}{Out-of-distribution accuracy endpoints and paired changes (pp). Intervals are descriptive joint run/item intervals.}
\label{tab:role_ood_test}
\end{minipage}\par\medskip
\paragraph{Retention accuracy.} The retention head evaluates the fixed MMLU anchor.
\par\medskip\noindent\begin{minipage}{\columnwidth}
\centering\small\setlength{\tabcolsep}{2pt}
\begin{tabular}{lrrrr}
\toprule
Model & $G_1$ & $G_2$ & $\Delta$ & 95\% CI\\
\midrule
Qwen4B & $-0.66$ & $-1.03$ & $-0.37$ & $[-0.98,+0.16]$\\
Llama3B & $+0.68$ & $+0.68$ & $-0.01$ & $[-1.14,+1.14]$\\
Smol3B & $+0.09$ & $-0.24$ & $-0.33$ & $[-1.35,+0.64]$\\
Qwen8B & $-1.03$ & $-1.59$ & $-0.56$ & $[-1.75,+0.50]$\\
Llama8B & $+1.17$ & $+0.34$ & $-0.83$ & $[-2.20,+0.47]$\\
\bottomrule
\end{tabular}
\captionof{table}{Retention accuracy endpoints and paired changes (pp). Intervals are descriptive joint run/item intervals.}
\label{tab:role_retention_test}
\end{minipage}\par\medskip

\paragraph{Code task.}
Each model/task stratum uses its own fixed seed grid.
\par\medskip\noindent\begin{minipage}{\columnwidth}
\centering\small\setlength{\tabcolsep}{2pt}
\begin{tabular}{lrrrr}
\toprule
Model & $G_1$ & $G_2$ & $\Delta$ & 95\% CI\\
\midrule
Qwen4B & $+1.73$ & $+1.61$ & $-0.12$ & $[-1.13,+0.91]$\\
Llama3B & $+1.19$ & $-1.23$ & $-2.42$ & $[-4.07,-0.86]$\\
Smol3B & $-4.74$ & $-9.60$ & $-4.87$ & $[-6.80,-3.12]$\\
Qwen8B & $-0.16$ & $-0.77$ & $-0.61$ & $[-1.81,+0.58]$\\
Llama8B & $-3.94$ & $-5.64$ & $-1.70$ & $[-3.82,+0.24]$\\
\bottomrule
\end{tabular}
\captionof{table}{Code endpoints and paired continuation changes (pp). There are eight runs per smaller-model stratum and four per 8B stratum.}
\label{tab:task_code}
\end{minipage}\par\medskip
\paragraph{Math task.}
Each model/task stratum uses its own fixed seed grid.
\par\medskip\noindent\begin{minipage}{\columnwidth}
\centering\small\setlength{\tabcolsep}{2pt}
\begin{tabular}{lrrrr}
\toprule
Model & $G_1$ & $G_2$ & $\Delta$ & 95\% CI\\
\midrule
Qwen4B & $-2.23$ & $-2.33$ & $-0.10$ & $[-1.42,+1.34]$\\
Llama3B & $-0.62$ & $-5.09$ & $-4.47$ & $[-6.58,-2.46]$\\
Smol3B & $-2.08$ & $-5.73$ & $-3.66$ & $[-6.98,-0.76]$\\
Qwen8B & $-0.72$ & $-1.47$ & $-0.75$ & $[-2.69,+1.15]$\\
Llama8B & $-2.90$ & $-6.20$ & $-3.30$ & $[-5.81,-0.67]$\\
\bottomrule
\end{tabular}
\captionof{table}{Math endpoints and paired continuation changes (pp). There are eight runs per smaller-model stratum and four per 8B stratum.}
\label{tab:task_math}
\end{minipage}\par\medskip

\section{Matched Control Results}
\label{app:controltable}

The recipient contrasts compare update increments from $M_1$ and $M_0$ on
fixed donor corrections. $D_0$ denotes corrections generated by the unchanged
parent on the control runtime; $D_1$ denotes archived candidate-generated
corrections. Their interaction is the difference of these receiver contrasts.
All paired receivers and native baselines are scored on the control runtime.
The reference-source comparison matches training problems and receiving
states, but not solution style or assistant-label token counts.
Recorded completion lengths are descriptive; no missing finish reason is
reconstructed as a known truncation event.

\par\medskip\noindent\begin{minipage}{\columnwidth}
\centering\small\setlength{\tabcolsep}{2pt}
\begin{tabular}{lrr}
\toprule
Contrast & Mean & Simultaneous 95\% CI\\
\midrule
Native first step & $-0.44$ & $[-2.40,+1.36]$\\
Native continuation & $-1.20$ & $[-3.06,+0.55]$\\
10 vs. 50 steps & $+0.47$ & $[-0.90,+1.94]$\\
25 vs. 50 steps & $-0.28$ & $[-1.37,+0.83]$\\
Pooled vs. two fixes & $+0.29$ & $[-1.57,+2.04]$\\
Recipient, $D_0$ & $-0.47$ & $[-3.02,+2.03]$\\
Recipient, $D_1$ & $-1.10$ & $[-3.85,+1.72]$\\
Recipient $\times$ donor & $-0.63$ & $[-2.79,+1.56]$\\
Gold vs. self-generated & $-11.56$ & $[-14.45,-8.64]$\\
\bottomrule
\end{tabular}
\captionof{table}{The nine declared controls (pp). Positive recipe contrasts favor the alternative over its matched baseline.}
\label{tab:controls}
\end{minipage}\par\medskip
Holm-adjusted $p=0.0018$ for gold versus self-generated corrections, $p=0.5407$ for the native continuation increment, and $p=1.0$ for the other seven contrasts.

\section{Selection Optimism Diagnostic}
\label{app:selection}

This corrected-template diagnostic evaluates candidate selection under
repeated continuation outcomes. Choosing and scoring a candidate on the
same draw yields optimism of $0$, $1.25$, and $2.09\pp$ for $K=1,2,3$.
The corresponding cross-draw selection gaps are $0$, $1.20$, and $2.25\pp$.
Synthetic null and recoverable-signal controls check that the analysis can
distinguish noise-only selection from a usable signal.
These results motivate independent selection and scoring; they are not
additional independent evidence for the fresh continuation contrast.

\par\medskip\noindent\begin{minipage}{\columnwidth}
\centering\includegraphics[width=\columnwidth]{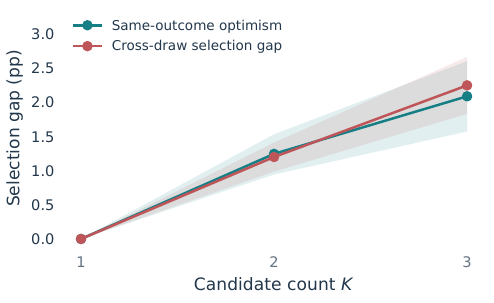}
\captionof{figure}{Selection gaps in an eight-parent diagnostic.
Intervals resample parents conditional on its scored items.
This diagnostic is separate from the fresh five-model estimates.}
\label{fig:optimism}
\end{minipage}\par\medskip

\section{Output Format and Length Audit}
\label{app:format}

The audit reads saved completions and reuses the scoring parser.
GSM8K failures mean that no unambiguous numeric answer could be extracted.
Code failures mean that the extracted program, including HumanEval's
required prefix where applicable, fails Python syntax parsing.
No code is executed again and no answer is repaired.
MATH's symbolic extraction is not assigned a comparable diagnostic rate.
Lengths count whitespace-separated words, rather than model tokens.
Generation finish reasons were not recorded, so these observations cannot
establish which outputs were truncated at the generation limit.

\par\medskip\noindent\begin{minipage}{\columnwidth}
\centering\small\setlength{\tabcolsep}{2pt}
\begin{tabular}{llrrr}
\toprule
Model & Head & Parent & $H_1$ & $H_2$\\
\midrule
Qwen4B & Code T & 0.00 & 0.83 & 1.01\\
Qwen4B & Code O & 2.34 & 1.56 & 1.56\\
Qwen4B & Math T & 4.69 & 3.22 & 4.82\\
Llama3B & Code T & 1.56 & 1.95 & 2.10\\
Llama3B & Code O & 0.78 & 1.22 & 1.79\\
Llama3B & Math T & 2.34 & 4.49 & 10.19\\
Smol3B & Code T & 0.00 & 1.95 & 5.50\\
Smol3B & Code O & 0.78 & 2.54 & 6.22\\
Smol3B & Math T & 3.12 & 4.69 & 15.12\\
Qwen8B & Code T & 0.00 & 0.00 & 0.00\\
Qwen8B & Code O & 1.56 & 1.56 & 1.43\\
Qwen8B & Math T & 8.59 & 5.27 & 6.02\\
Llama8B & Code T & 2.34 & 0.20 & 0.85\\
Llama8B & Code O & 0.78 & 1.37 & 1.24\\
Llama8B & Math T & 2.34 & 2.73 & 7.55\\
\bottomrule
\end{tabular}
\captionof{table}{Syntax or numeric-extraction failure rates (\%). T denotes target and O denotes OOD.}
\label{tab:format}
\end{minipage}\par\medskip
The same development-selected candidate weights are used for these format rates and for utility.
\paragraph{Completion length.} Word counts characterize output changes without inferring token-level truncation.
\par\medskip\noindent\begin{minipage}{\columnwidth}
\centering\small\setlength{\tabcolsep}{2pt}
\begin{tabular}{llrr}
\toprule
Model & Head & $H_1$ words & $H_2$ words\\
\midrule
Qwen4B & Code T & 44.3 & 47.0\\
Qwen4B & Code O & 124.8 & 123.7\\
Qwen4B & Math T & 147.8 & 144.9\\
Llama3B & Code T & 73.4 & 85.5\\
Llama3B & Code O & 149.9 & 141.9\\
Llama3B & Math T & 149.2 & 164.1\\
Smol3B & Code T & 130.9 & 159.8\\
Smol3B & Code O & 185.5 & 190.1\\
Smol3B & Math T & 147.4 & 187.6\\
Qwen8B & Code T & 25.8 & 28.0\\
Qwen8B & Code O & 146.2 & 139.4\\
Qwen8B & Math T & 156.4 & 169.3\\
Llama8B & Code T & 61.0 & 59.1\\
Llama8B & Code O & 99.9 & 91.9\\
Llama8B & Math T & 144.5 & 161.0\\
\bottomrule
\end{tabular}
\captionof{table}{Mean whitespace-separated completion lengths. These are descriptive output lengths, not generated-token counts.}
\label{tab:length}
\end{minipage}\par\medskip

\section{Statistical and Artifact Accounting}

The main five-model analysis contains 64 fresh runs, 192 immediate candidates,
576 continuations, and 832 scored checkpoints. Each checkpoint has 128 target,
128 OOD, and 256 retention outcomes. Role weights are equal irrespective of
item count. The 16 control parents overlap previously inspected runs and
are reported separately.

Bootstrap resampling preserves parent strata and shared item identities.
The three native draws are averaged before bootstrap resampling.
Consequently, neither checkpoints nor native draws are used as independent
parent observations. Primary reports retain their declared multiplicity
families; the five-model plotting analysis is a separate descriptive
re-expression of saved outcomes.

The accompanying source includes vector figures, aggregate plotting data,
analysis code, and manifest/report checksums.
The saved summaries reproduce the figures without further model execution.

\end{document}